\documentclass[10pt,twocolumn,letterpaper]{article}

\usepackage[pagenumbers]{cvpr} 

\usepackage[dvipsnames]{xcolor}

\usepackage{multirow}
\usepackage{colortbl}
\usepackage{booktabs}
\usepackage{bbding}

\definecolor{mygray}{gray}{.92}

\definecolor{cvprblue}{rgb}{0.21,0.49,0.74}
\usepackage[pagebackref,breaklinks,colorlinks,citecolor=cvprblue]{hyperref}

\def\paperID{*****} 
\def\confName{CVPR}
\def\confYear{2024}

\title{Rolling Shutter Correction with Intermediate Distortion Flow Estimation}

\author{
Mingdeng~Cao$^1$ \quad
Sidi~Yang$^2$ \quad
Yujiu~Yang$^2$ \quad
Yinqiang~Zheng$^1$ \\
$^1$The University of Tokyo \quad
$^2$Tsinghua University \\
}

\begin{document}
\maketitle
\begin{abstract}
This paper proposes to correct the rolling shutter (RS) distorted images by estimating the distortion flow from the global shutter (GS) to RS directly. Existing methods usually perform correction using the undistortion flow from the RS to GS. They initially predict the flow from consecutive RS frames, subsequently rescaling it as the displacement fields from the RS frame to the underlying GS image using time-dependent scaling factors. Following this, RS-aware forward warping is employed to convert the RS image into its GS counterpart. Nevertheless, this strategy is prone to two shortcomings. First, the undistortion flow estimation is rendered inaccurate by merely linear scaling the flow, due to the complex non-linear motion nature. Second, RS-aware forward warping often results in unavoidable artifacts. To address these limitations, we introduce a new framework that directly estimates the distortion flow and rectifies the RS image with the backward warping operation. More specifically, we first propose a global correlation-based flow attention mechanism to estimate the initial distortion flow and GS feature jointly, which are then refined by the following coarse-to-fine decoder layers.
Additionally, a multi-distortion flow prediction strategy is integrated to mitigate the issue of inaccurate flow estimation further. Experimental results validate the effectiveness of the proposed method, which outperforms state-of-the-art approaches on various benchmarks while maintaining high efficiency. The project is available at \url{https://github.com/ljzycmd/DFRSC}.
\end{abstract}
\section{Introduction}
\label{sec:intro}

\begin{figure}
    \centering
    \includegraphics[width=\linewidth]{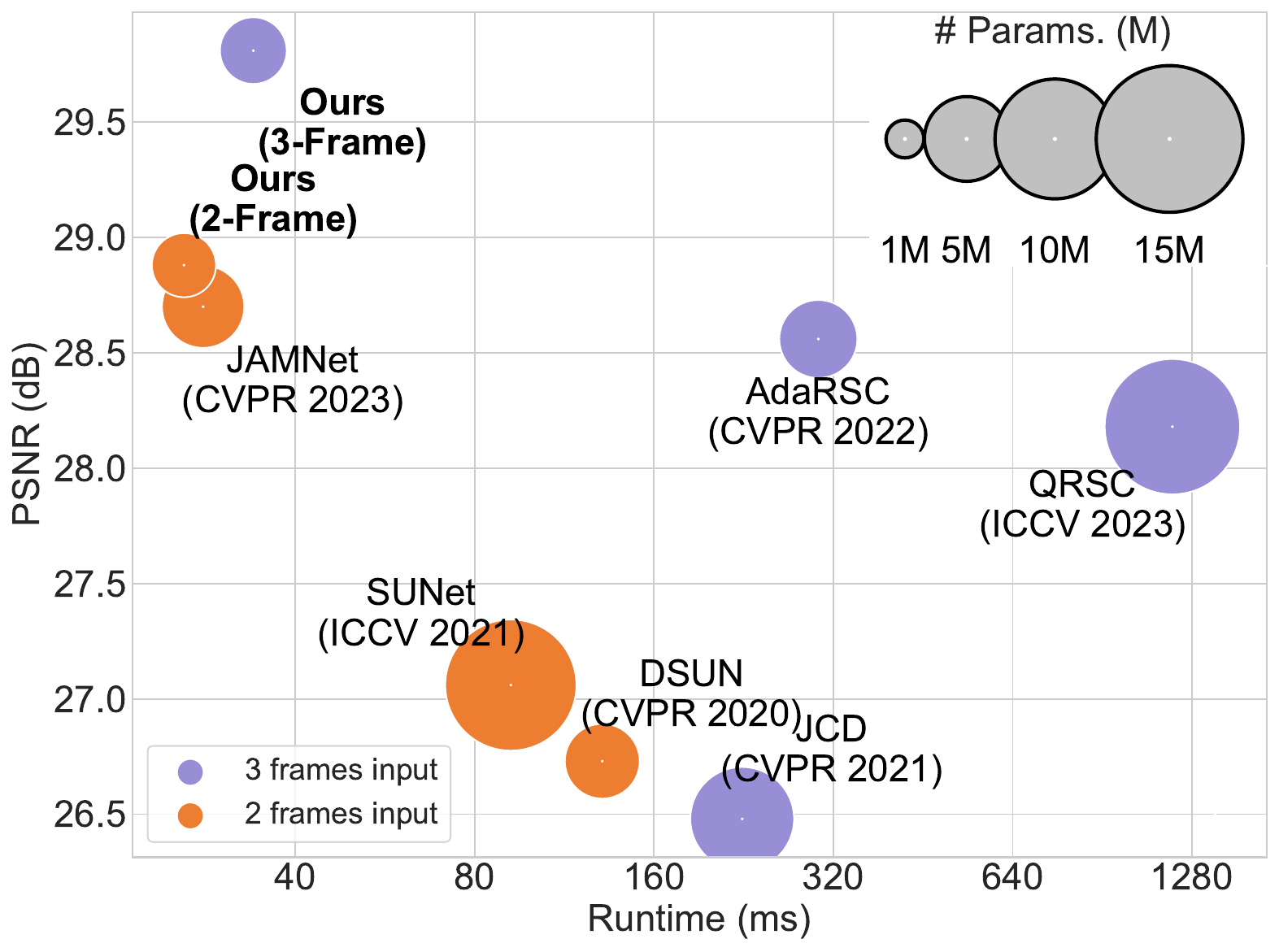}
    \caption{Model comparison in terms of PSNR (dB), runtime, and model size. The PSNR and runtime are calculated on the Fastec-RS~\cite{liu2020deep} dataset with a resolution of $640 \times 480$ using an RTX 3090 GPU. The proposed method outperforms the state-of-the-art rolling shutter correction methods with higher efficiency. }
    \label{fig:teaser}
    \vspace{-3mm}
\end{figure}

We often encounter distorted images/videos when relative movements occur between the scene and the camera during the acquisition process. For instance, a straight building may appear slanted in the captured photograph, while the blades of a flying helicopter may seem distorted. This phenomenon is generally referred to as the wobble or the “Jello effect.”,  which is caused by the rolling shutter (RS) mechanism of cameras. The image pixels are exposed from the top to the bottom sequentially, instead of capturing the entire frame all at once as in a global shutter (GS) camera. This RS mechanism is employed in CMOS sensors, which govern the cameras~(\eg, smartphones, digital cameras) in the consumer market, owing to their fast imaging and low cost. However, some unintended distortions would occur in the image content when capturing moving scenes, affecting our visual perception and deteriorating the performance of downstream tasks, especially 3D vision tasks~\cite{dai2016rolling, kim2016direct, engel2017direct, liao2023revisiting}. Consequently, developing effective and robust image/video RS correction (RSC) algorithms to remove such distortions holds significant research and practical application value.

To recover the latent distortion-free GS image corresponding to a specific exposure scanline of the RS image, previous research efforts~\cite{albl2020two, fan2021rolling, rengarajan2016bows, purkait2017rolling} have attempted to directly restore the underlying GS image from a single RS image by employing additional geometric constraints and priors. However, this kind of single-image-based approach is highly challenging and has limited effectiveness, as the motion states that form the distortion are unknown and strongly ambiguous, making the removal of such distortions from a single image highly ill-posed. Utilizing multiple consecutive images can significantly alleviate this issue by extracting the inter-frame motion information, thereby achieving better and more robust results~\cite{grundmann2012calibration, zhuang2017rolling, lao2018robust, vasu2018occlusion, zhuang2020image}, especially with deep-learning-based techniques~\cite{liu2020deep, fan2021sunet, cao2022learning, fan2023joint, qu2023towards}. Generally, to obtain the displacement fields (\ie, the correction fields) from RS to GS images, these methods usually first estimate the inter-frame motion field (optical flow) between the RS frames and then use the relationship~\cite{zhuang2017rolling, fan2021inverting, fan2022context, qu2023towards} between RS and GS images to transform it into a correction field, thus obtaining the desired GS images through image warping operations. These methods have achieved great success but struggle when faced with non-linear and large motion due to the following reasons:
\textbf{1) First,} some methods~\cite{fan2021inverting, fan2022context, naor2022combining, qu2023towards} employ off-the-shelf motion modeling networks to estimate the inter-frame motion of RS images. However, since these optical flow estimation networks have not been trained on RS videos, their estimation results may contain distortions and exhibit erroneous dynamic behavior, making it difficult to obtain an accurate correction field for recovering underlying GS images.
\textbf{2) Alternatively}, other methods~\cite{liu2020deep, zhong2021towards, fan2021sunet, cao2022learning} estimate the inter-frame optical flow within the RSC model and are trained with RS images/videos, typically using local correlation~\cite{dosovitskiy2015flownet}, which makes it difficult to model large motions.
\textbf{3) Moreover}, to obtain the correction field, the estimated inter-frame optical flow needs to be further linearly scaled based on the constant velocity assumption~\cite{liu2020deep, cao2022learning, fan2023joint}. However, the motion in the real world is highly non-linear, rendering the obtained correction field inaccurate. Although the recent work~\cite{qu2023towards} proposes estimating a quadratic correction field, the motion in real-world scenarios is often more complex than quadratic and is thus more challenging to model.

To move beyond these limitations, in this paper, we propose to directly estimate the correction field from GS to RS images, dubbed \textit{Distortion Flow}~\footnote{Distinguished with the \textit{undistortion flow} field from RS to GS, used in~\cite{liu2020deep, zhong2021towards, fan2021sunet, fan2021inverting, fan2022context, qu2023towards}.}. 
More specifically, we first generate the latent GS feature based on the extracted RS features. Then, we obtain an initial estimation of the distortion flow by establishing the global correlation between the RS and GS features, with the proposed flow attention mechanism. The GS feature and flow are continuously refined through a coarse-to-fine decoder, which fuses the warped RS appearance information to the GS feature and updates the flow. Simultaneously, we integrate a multi-distortion field decoding strategy to further alleviate the occlusion problem. The RS features are backwardly warped into the GS counterparts using multiple distortion fields and decoded along with the GS features to generate the final GS image. As shown in Fig.~\ref{fig:teaser}, our method achieves highly competitive results on various datasets more efficiently.

Our contributions are threefold and can be summarized as follows.
\textbf{1)} We propose a novel framework for the RSC task that directly predicts \textit{Distortion Flow} from consecutive RS frames to recover the underlying GS frame.
\textbf{2)} We design a global correlation-based flow attention mechanism for GS feature and flow prediction, facilitating large motion prediction. In conjunction, a multi-distortion flow prediction strategy is formulated to further improve the performance. 
\textbf{3)} Extensive experiments demonstrate that the proposed method achieves substantial performance improvements against state-of-the-art methods on multiple datasets while maintaining higher efficiency.

\section{Related Work}
\label{sec:related_work}

\subsection{Deep Rolling Shutter Correction}
Existing works of RSC fall into two categories: single-image-based and multi-frame-based methods. For the former, previous methods apply different geometric assumptions, such as straight lines kept straight~\cite{Rengarajan_2016_CVPR}, vanishing direction restraint~\cite{purkait2017rolling}, and analytical 3D straight line RS projection model~\cite{lao2018robust}. Driven by the surge of deep learning, the first learning-based model proposed in~\cite{Rengarajan_2017_CVPR} attempts to remove RS distortions from a single distorted image. However, single-image-based models often exhibit unsatisfactory performance due to their reliance on either strong assumptions or inconspicuous features. This limitation hinders their ability to accurately capture the complexity of the underlying data, leading to sub-optimal results.

To tackle these limitations, multi-frame-based methods are adopted to model the RS motion, which can be categorized into classical and learning-based models. For the classical methods, modeling the RS motion from the uncalibrated RS images and two consecutive frames are respectively studied in \cite{grundmann2012calibration, Lao_2018_CVPR} and~\cite{vasu2018occlusion, zhuang2020image}. For the learning-based methods, the works~\cite{liu2020deep, fan2021sunet} are proposed to model the RS motions between two consecutive RS frames by constructing cost volumes. Considering the blur in RS images, Zhong~\etal~\cite{zhong2021towards} further designed a three-frame-based model to remove the blur and RS distortion simultaneously. To alleviate the inaccurate displacement field estimation and warping, Cao~\etal~\cite{cao2022learning} proposed to predict multiple fields and warp the RS features adaptively. Fan~\etal\cite{fan2023joint} and Qu~\etal\cite{qu2023towards} proposed a joint motion and appearance modeling network and a quadratic RS motion solver, respectively, achieving new heights.

\subsection{Inter-frame Motion Modeling}
To model the motions across frames, computing the matching cost volume to obtain the correspondence is a classical way. Optical flow networks, such as~\cite{sun2018pwc, teed2020raft, ilg2017flownet, dosovitskiy2015flownet}, usually apply local correlations to obtain the final flow in a coarse-to-fine strategy with efficiency. RAFT~\cite{teed2020raft} proposes an all-pair correlation volume and designs a recurrent strategy to refine the predicted flow continuously, obtaining considerable accuracy improvement. GMFlow~\cite{xu2022gmflow} also constructs a global correlation between two frames to aggregate the coordinate grid as the correspondence, forming a new paradigm for optical flow estimation.  These models are widely employed as off-the-shelf modules for motion modeling in many video-related tasks, like video frame interpolation, video enhancement, video editing, and RSC task~\cite{fan2021inverting, fan2022context, naor2022combining, qu2023towards}. However, since these methods are not trained in the specific domain data, the motion estimation in the other tasks is inaccurate. To this point, Super-Slomo~\cite{jiang2018super} introduces a mask to handle the occlusion explicitly and provides a standard formulation for synthesizing intermediate frames. RIFE~\cite{huang2022real} and IFRNet~\cite{kong2022ifrnet} propose task-oriented flow distillation losses to provide a prior intermediate flow in training. AMT~\cite{li2023amt} further adapts the all-pair correlation for efficient frame interpolation. In this paper, we also propose to learn the distortion flow for the RSC task, by constructing global correlations between the underlying GS image and input RS frames. 
\section{Proposed Method}

\subsection{Preliminary}

\begin{figure}
    \centering
    \includegraphics[width=\linewidth]{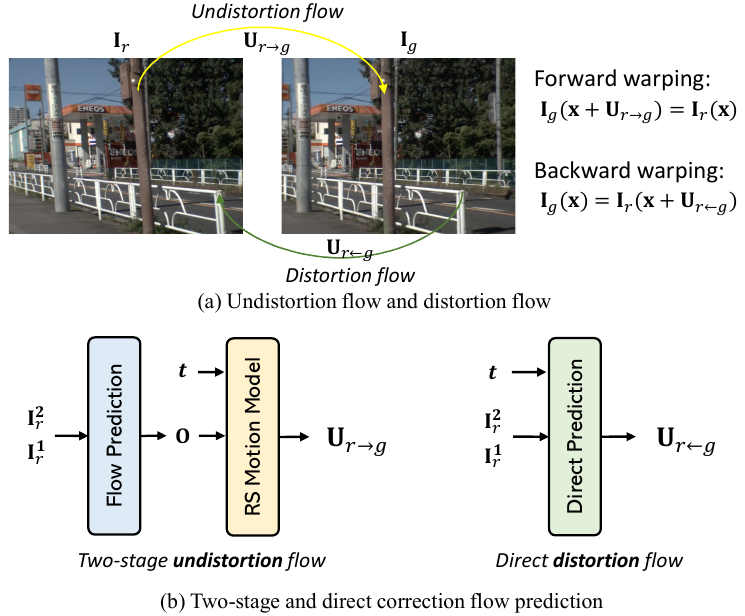}
    \caption{(a) Schematic of the undistortion flow and distortion flow. (b) Comparison between the two-stage flow estimation and the proposed direct distortion flow estimation.}
    \label{fig:distortion_flow}
\end{figure}

RS cameras expose the pixels in a row-by-row manner, and each scanline has a different timestamp and motion. Thus the RS image $\mathbf{I}_r$ can be formed by row-by-row stacking the virtual GS images corresponding to each row timestamp:
\begin{equation}
    \label{eq:rs_imaging}
    [\mathbf{I}_r(\mathbf{x})]_i = [\mathbf{I}^i_g(\mathbf{x})]_i,\quad 0\le i \le H-1,
\end{equation}
where $\mathbf{I}^i_g$ is the virtual GS image corresponding to the timestamp of $i$-th RS image row, $[\cdot]_i$ is the operation to extract the $i$-th image row,  $H$ and $\mathbf{x}$ are respectively the image height and the pixel location. More generally, we can obtain the $j$-th RS image row with the displacement field $\mathbf{u}_{i\rightarrow j}$ from the $j$-th row of the RS image to the $i$-th virtual GS image:
\begin{equation}
    \label{eq:gs2rs}
    [\mathbf{I}_r(\mathbf{x})]_j = [\mathbf{I}^i_g(\mathbf{x}+\mathbf{u}_{j\rightarrow i})]_j, \quad 0\le i, j \le H-1.
\end{equation}
With the above equation, we can obtain the \textit{RS undistortion flow} field (the yellow line in Fig.~\ref{fig:distortion_flow}(a)) $\textbf{U}_{r\rightarrow i}$ from the RS image to the $i$-th virtual GS image, by stacking all $\mathbf{u}_{j\rightarrow i}$ from $j=0$ to $j=H-1$. Therefore, we can recover the $i$-th underlying GS image~(usually corresponding to the first scanline~\cite{fan2021sunet}, and the middle scanline~\cite{liu2020deep, cao2022learning, fan2023joint, qu2023towards} of the RS image) by estimating the undistortion flow field and using a forward warping operation like the differential forward warping (DFW) module~\cite{liu2020deep}. As shown in the left part of Fig.~\ref{fig:distortion_flow}(b), the velocity of the RS image pixels (approximated as the optical flow between consecutive RS frames) is first estimated, then the undistortion flow can be usually calculated by rescaling the flow under the constant velocity assumption~\cite{liu2020deep, fan2021inverting, fan2022context, cao2022learning, fan2023joint}. However, accurate $\textbf{U}_{r\rightarrow i}$ is hard to estimate with such a linear model since the motion in the real world is highly complex and non-linear, even with a recently proposed quadratic motion solver~\cite{qu2023towards}. Moreover, the inaccurate $\mathbf{U}_{r\rightarrow i}$ further results in undesired warping artifacts with the DFW module, \eg, black holes shown in~\cite{fan2021inverting}.

In contrast to the Eq.~\ref{eq:gs2rs} that forms the RS image from a sequence of GS images, it is feasible to derive the underlying $i$-th GS image from an RS image, when with the motion displacement field $\mathbf{U}_{r\leftarrow i}$ from GS to RS images (the green line in Fig.~\ref{fig:distortion_flow}(a)):
\begin{equation}
    \label{eq:rs2gs}
    \mathbf{I}^i_g(\textbf{x}) = \mathbf{I}_r(\mathbf{x} + \mathbf{U}_{r\leftarrow i}(\mathbf{x})).
\end{equation}
Thus we can obtain the underlying GS image by sampling pixels in the RS image with interpolation operations, \eg, bilinear, and bicubic. Since the motion field attributes distort the GS image into the RS image, we dub it \textit{distortion flow} field. However, the challenge arises since the intermediate GS image is unknown. 

In this work, we propose to estimate the intermediate distortion flow from the underlying desired GS image to the RS image in a single-stage manner, as depicted in the right part of Fig.~\ref{fig:distortion_flow}(b). Note that the time $\mathbf{t}$ determines the recovery of a specific GS image, and is optional for the RSC task since we aim to recover only one GS frame corresponding to a specific scanline (first or middle) of the RS frame. 

\begin{figure*}[!t]
    \centering
    \includegraphics[width=\linewidth]{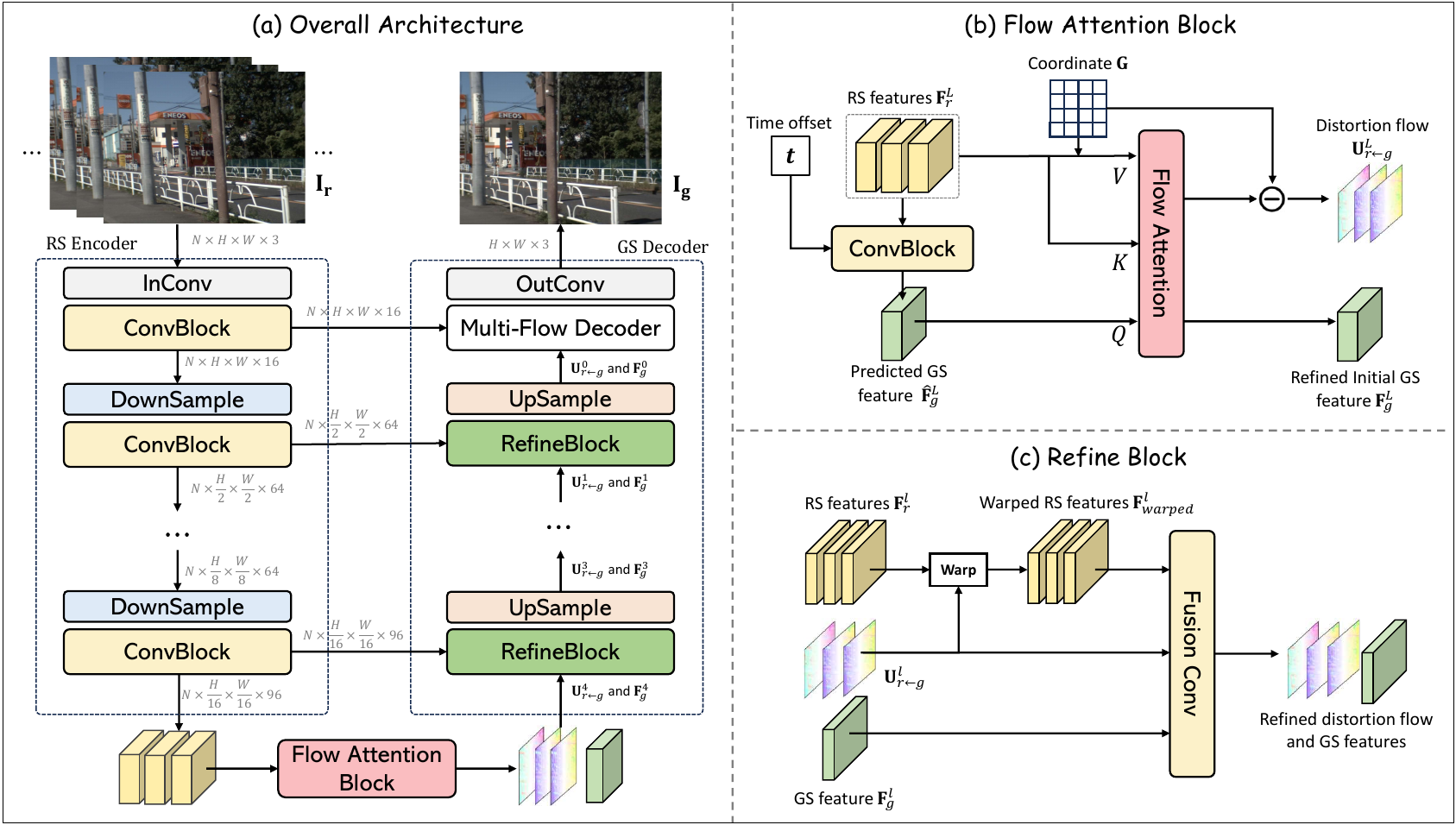}
    \caption{Overview of the proposed method (a) and the detailed architecture of the key components (b), (c). Our model directly predicts the distortion flow for efficient and high-quality RSC. }
    \label{fig:model_arch}
\end{figure*}

\subsection{Model Overview}
Our method aims to alleviate the inaccurate motion modeling under large and complex non-linear motions in the RSC task, by directly estimating the intermediate distortion flows. Our method takes $N$ consecutive RS frames as input, and recovers the latent GS image corresponding to the timestamp of the middle scanline of the middle input RS frame, consistent with the settings in previous works~\cite{liu2020deep, zhong2021towards, cao2022learning, fan2023joint, qu2023towards}). The overall architecture of the proposed method is illustrated in Fig.~\ref{fig:model_arch}. We first extract multi-scale frame-level RS features, using a weight-sharing image encoder. After that, we obtain the initial distortion flow along with the GS features at the lowest resolution with a global correlation-based flow attention mechanism. Then, the coarse-to-fine decoder refines and upscales the resolution of the flow and GS features simultaneously. The final GS image is obtained by a multi-flow predicting strategy.

\subsection{Intermediate Distortion Flow Estimation}
\vspace{2mm}
\noindent\textbf{Initial distortion flow estimation.}
After obtaining the $L$-scale features $\{\mathbf{F}^{l}\}^{L}_{l=0}$ of the input $N$ RS frames $\mathbf{I}_r\in \mathbb{R}^{N \times H\times W \times 3}$ from the encoder, we can directly predict the initial intermediate distortion flow $\mathbf{U}_{r\leftarrow g}$ with the lowest resolution features $\mathbf{F}^L\in \mathbb{R}^{N\times H' \times W' \times D}$ in a naive way:
\begin{equation}
    \label{eq:idfe}
    \mathbf{U}^{L}_{r\leftarrow g} = \text{IDFE}(\mathbf{F}^{L}, \textbf{t}),
\end{equation}
where IDFE is the prediction network, and $\mathbf{t}$ is the exposure time offset between the target GS image and the middle scanline of the RS frame.

To obtain a more accurate intermediate distortion flow estimation under large motions, we further perform global correlation modeling across the desired underlying GS and RS features. The $\text{Attention}=\text{Softmax}(\frac{\mathbf{QK}^T}{\sqrt{d}})\mathbf{V}$ mechanism~\cite{vaswani2017attention} tries to aggregate the value $V \in \mathbb{R}^{S\times d}$ with the correlation between the query $Q \in \mathbb{R}^{S\times d}$ and value $K \in \mathbb{R}^{S\times d}$, excelling at long-range modeling and correlation modeling. We extend such an operation to build the global correlation between GS and RS frames for distortion flow estimation and RS feature warping shown in Fig.~\ref{fig:model_arch}(b), dubbed flow attention. While the GS feature is missing, we thus estimate it $\mathbf{F}^L_g\in \mathbb{R}^{H' \times W' \times D}$ firstly by fusing the consecutive RS features with the time offsets condition:
\begin{equation}
    \label{eq:gs_pred}
    \mathbf{F}^L_g = \text{ConvBlock}(\mathbf{F}^{L}, \textbf{t}).
\end{equation}
Let the GS features and RS features serve as \textit{query} and \textit{key}, respectively, and we can compute the attention map between them:
\begin{equation}
    \label{eq:correlation}
    \mathbf{M} = \text{Softmax}\left(\frac{\mathbf{F}^L_g\mathbf{F}^L}{\sqrt{D}}\right) \in \mathbb{R}^{N\times H' \times W' \times H' \times W'},
\end{equation}
where each element $(n, i, j, k, l)$ in $\mathbf{M}$ represents the correspondence probability between the GS feature $\mathbf{F}^L_g(i, j)$ and the RS frame feature $\mathbf{F}^{L}(n, k, l)$. Note that the above equation is consistent with the differentiable matching layer~\cite{wang2020learning, xu2022gmflow} in image matching and optical flow estimation. With the global correlation matrix $\mathbf{M}$, we can simultaneously compute the globally warped RS features and the distortion flow by aggregating 1) the RS features and 2) the 2D coordinates grid $\mathbf{G}\in \mathbb{R}^{H' \times W' \times 2}$ of the RS frame, respectively. As a result, both $\textbf{F}^L$ and $\textbf{G}$ serve as the \textit{value} for the decoding:
\begin{align}
    \label{eq:flow_pred}
    \mathbf{F}^L_{warped} = \mathbf{M}\mathbf{F}^L \in \mathbb{R}^{N\times H' \times W' \times D}, \\
    \mathbf{U}^L_{r\leftarrow g} = \mathbf{MG} - \mathbf{G} \in \mathbb{R}^{N\times H' \times W' \times 2}.
\end{align}
The warped RS features are further used to refine the predicted $\mathbf{F}^L_g$ using a convolutional block.

Employing the global correlation attention mechanism enables obtaining a more precise distortion flow from RS frames, especially with non-linear and large motions. In addition, the predicted GS feature can be further refined with the globally warped RS features, by fusing the complementary information RS features.

\vspace{2mm}
\noindent\textbf{Progressive refinement.}
The predicted initial distortion flow and GS feature at the lowest resolution are progressively refined by the decoder. Inspired by~\cite{fan2023joint}, we employ the joint appearance and motion refinement strategy, while we directly predict the upsampled refined distortion flow rather than scale the optical flow between RS frames. Specifically, given the current refined distortion flow $\textbf{U}^l_{r\leftarrow g}$ and the GS features $\textbf{G}^l_g$ at level $l$, we first warp the RS features extracted from the image encoder at the corresponding level to the GS candidates:
\begin{equation}
    \label{eq:decoder_rs_warping}
    \mathbf{F}^l_{warped} = \mathcal{W}(\mathbf{F}^l, \textbf{U}^l_{r\leftarrow g}),
\end{equation}
where $\mathcal{W}$ is the backward warping operation.  Next, the warped RS features, distortion flow, and GS feature are fused and upscaled  to the refined distortion flow and GS feature at the next scale $l-1$:
\begin{equation}
    \label{eq:decoder_stage}
    \textbf{U}^{l-1}_{r\leftarrow g}, \textbf{F}^{l-1}_g = \text{Upsample}(\text{FusionBlock}(\textbf{U}^{l}_{r\leftarrow g}, \textbf{F}^{l}_g, \mathbf{F}^l_{warped})).
\end{equation}
By progressively fusing the complementary information from the RS features, more accurate distortion flow and corresponding GS features are obtained to be decoded as the final GS image.

\vspace{2mm}
\noindent\textbf{Multi-distortion flow fields decoding.}
 At the 0-level with the largest resolution, we further employ a multiple distortion fields prediction strategy~\cite{cao2022learning} to alleviate some incorrectly estimated displacement in the distortion flow.  With the refined $\mathbf{U}^{0}_{r\leftarrow g}$ and $\mathbf{F}^0_g$, rather than utilize them to synthesize the GS image directly, we instead predict multiple groups of fields with a convolutional block:
\begin{equation}
     \label{eq:multi_flows}
     \{\mathbf{U}_{1, r\leftarrow g}, \cdots, \mathbf{U}_{G, r\leftarrow g}\} = \text{ConvBlock}(\textbf{U}_{r\leftarrow g}, \textbf{F}_g, \mathbf{F}_{warped})).
\end{equation}
Therefore, the RS features $\mathbf{F}^{0}$ are further warped according to Eq.~\ref{eq:decoder_rs_warping}, resulting in $G$ groups of warped RS features. The final GS image $\mathbf{I}_g$ is predicted from the refined GS features, undistortion flow, and warped multiple RS features.

\subsection{Training Strategy}

\noindent\textbf{GS image supervision. }
Following previous works~\cite{liu2020deep, fan2021sunet, zhong2021towards, cao2022learning, fan2023joint}, we employ a combination of the Charbonnier loss~\cite{lai2017deep} 
\begin{equation}
    \label{eq:charbonnier_loss}
    \mathcal{L}_c = d(\mathbf{I}_{gt}-\mathbf{I}_{g})
\end{equation}
and the Perceptual loss~\cite{johnson2016perceptual}
\begin{equation}
    \label{eq:perceptual_loss}
    \mathcal{L}_p = \Vert \phi(\mathbf{I}_{gt}) - \phi(\mathbf{I}_{g}) \Vert_1,
\end{equation}
for the recovered GS image supervision, where $d(x) = \sqrt{(x)^2 + \epsilon^2}$ is a distance function, and $\epsilon$ is set to $1e^{-3}$, and $\phi$ is the extractor to obtain the features from layer \texttt{Conv5\_4} of pretrained VGG-19 network~\cite{simonyan2014very}.

\vspace{2mm}
\noindent\textbf{Distortion flow supervision. }
To ensure the accuracy of the estimated distortion flow, we employ an indirect supervision method that ensures the backward warped RS images with the undistortion flow align consistently with the ground truth GS image:
\begin{equation}
    \label{eq:consistency_loss}
    \mathcal{L}_{w} = \frac{1}{L}\sum^{L}_{l=0} d(\mathbf{I}^{l, warped}_{r} - \mathbf{I}^{l}_{gt}),
\end{equation}
where $\mathbf{I}^{l, warped}_{r} = \mathcal{W}(\mathbf{I}^{l}_{r}, \mathbf{U}^l_{r\leftarrow g})$ is the warped downsampled RS frames $\mathbf{I}_r$ at the $l$-th level, and $\mathbf{I}^{l}_{gt}$ is the downsampled ground truth GS image at level $l$.

The total loss for the model training can be formulated as follows:
\begin{equation}
    \label{eq:loss}
    \mathcal{L} = \mathcal{L}_c + \lambda_1\mathcal{L}_p + \lambda_2\mathcal{L}_w,
\end{equation}
where $\lambda_1$ and $\lambda_2$ are loss weights. 
\section{Experiments}
\subsection{Experimental Setup}
\noindent\textbf{Datasets. }
We evaluate the proposed method on both synthetic datasets Fastec-RS~\cite{liu2020deep}, Carla-RS~\cite{liu2020deep}, and the real-world datasets BS-RSC~\cite{cao2022learning}. The Fastec-RS dataset is synthesized from the extremely high-speed videos captured by a GS camera, mainly containing RS effects caused by horizontal camera movements. Another synthetic dataset Carla-RS is generated from a virtual 3D environment, with constant translational velocity and angular rate during the RS video sequence generation process. The recently proposed BS-RSC dataset is collected from the real world. The RS videos and corresponding GS videos are captured simultaneously by a well-designed beam-splitter acquisition system. The scenes contain natural non-linear and large motions, including both camera and objects.

\begin{table*}[t]
    \centering
    \resizebox{\linewidth}{!}{
    \begin{tabular}{lccccccccc}
    \toprule
    \multirow{2}{*}{Method}     & \# Params. & Runtime & \# NF & \multicolumn{3}{c}{Fastec-RS} & \multicolumn{3}{c}{Carla-RS} \\ \cmidrule{5-7}\cmidrule{8-10}
        &  (Million) & (ms)  &   &  PSNR$\uparrow$(dB)  &  SSIM$\uparrow$  & LPIPS$\downarrow$ &  PSNR$\uparrow$(dB)  &  SSIM$\uparrow$  & LPIPS$\downarrow$ \\
    \midrule
    DiffSfM~\cite{zhuang2017rolling}       
        &  -  & $4.7\times10^{5}$ & 2 & 21.44 & 0.710 & 0.2180 & 21.28 & 0.775 & 0.1322 \\
    DSUN~\cite{liu2020deep}
        & 3.91 & 131 & 2 & 26.73 & 0.819 & 0.0995 & 26.46 & 0.807 & 0.0703 \\
    SUNet~\cite{fan2021sunet}
        & 12.0 & 92  & 2 & 27.06 & 0.825 & 0.1030 & 29.18 & 0.850 & 0.0658 \\
    VideoRS~\cite{naor2022combining}
        & 24.26 & $1.3\times10^{6}$ & 2 & 28.57 & 0.844 & - & \textbf{31.43} & \underline{0.919} & - \\
    JAMNet~\cite{fan2023joint}
        & 4.73 & 28  & 2 & \underline{28.70} & \underline{0.865} & \textbf{0.0691} & 30.70 & 0.905 & \underline{0.0371} \\
    \rowcolor{mygray}
    \textbf{Ours} (2F)
        &  \textbf{2.87} & \textbf{26} & 2 & \textbf{28.88} & \textbf{0.870} &  \underline{0.0699}  & \underline{31.33} & \textbf{0.921} & \textbf{0.0228} \\
    \midrule
    JCD~\cite{zhong2021towards}
        & 7.51 & 225 & 3 & 26.48 & 0.821 & 0.0943 & 27.75 & 0.836 & 0.0595 \\
    AdaRSC~\cite{cao2022learning}
        & 4.25 & 302 & 3 & 28.56 & 0.855 & \underline{0.0796} & - & - & - \\
    QRSC (3F)~\cite{qu2023towards}
        & 12.72 & 401 & 3 & 28.18 & 0.853 & 0.0912 & 29.81 & 0.919 & 0.0313 \\
    QRSC (4F)~\cite{qu2023towards}
        & 12.74 & 759 & 4 & 28.26 & 0.854 & 0.0901 & 30.98 & 0.925 & 0.0282  \\
    QRSC (5F)~\cite{qu2023towards}
        & 12.75 & 1149 & 5 & \underline{29.49} & 0.872 &  \underline{0.0814} & 32.01 & \textbf{0.933} & \underline{0.0253} \\
    \rowcolor{mygray}
    \textbf{Ours} (3F) 
        & \textbf{3.15} & \textbf{34} & 3 & \textbf{30.00} & \textbf{0.882} & \textbf{0.0665} & \textbf{32.10} & \underline{0.930} & \textbf{0.0218} \\
    \bottomrule
    \end{tabular}
    }
    \caption{Quantitative comparison against the state-of-the-art methods on the synthetic RSC datasets Carla-RS~\cite{liu2020deep} and Fastec-RS~\cite{liu2020deep}. Our method achieves highly competitive results while maintaining high efficiency. \#NF indicates the input RS frames of the model. The runtime is calculated using an NVIDIA RTX 3090 GPU.}
    \label{tab:quantitative_results}
\end{table*}

\begin{table}[h]
    \centering
    \resizebox{\linewidth}{!}{
    \begin{tabular}{lcccc}
    \toprule
    \multirow{2}{*}{Method}  &  \multicolumn{2}{c}{BS-RSC} & \multicolumn{2}{c}{ACC} \\ \cmidrule{2-3}\cmidrule{4-5}
        &  PSNR$\uparrow$(dB)  &  SSIM$\uparrow$  &  PSNR$\uparrow$(dB)  &  SSIM$\uparrow$ \\
    \midrule
    DiffSfm~\cite{zhuang2017rolling}
        & 19.80 & 0.698 & 15.74 & 0.551 \\
    DSUN~\cite{liu2020deep}
        & 25.21 & 0.833 & 22.39 & 0.780 \\
    SUNet~\cite{fan2021sunet}
        & 27.76 & 0.875 & 27.29 & 0.870 \\
    JAMNet~\cite{fan2023joint}
        & \underline{32.93} & \underline{0.941} & \underline{32.71} & \underline{0.940} \\
    \rowcolor{mygray}
    \textbf{Ours} (2F)
        &\textbf{33.39} & \textbf{0.947} & \textbf{33.21} & \textbf{0.947}\\
    \midrule
    JCD~\cite{zhong2021towards}
        & 25.59 & 0.841 & 23.73 & 0.808 \\
    AdaRSC~\cite{cao2022learning}
        & 28.23 & 0.882 & 28.73 & 0.892 \\
    QRSC (5F)~\cite{qu2023towards}
        & \underline{33.50} & \underline{0.946} & \underline{33.36} & \underline{0.945} \\
    \rowcolor{mygray}
    \textbf{Ours} (3F)
        & \textbf{34.48} & \textbf{0.954} & \textbf{34.35} & \textbf{0.954} \\
    \bottomrule
    \end{tabular}
    }
    \caption{Quantitative comparison against the state-of-the-art methods on the real-world RSC dataset BS-RSC~\cite{cao2022learning}. }
    \label{tab:quantitative_results_real}
    \vspace{-3mm}
\end{table}

\vspace{2mm}
\noindent\textbf{Implementation details. }
During the training process, our model accepts $N=3$ consecutive RS frames in RGB format as input, while we also train a 2-frame-based model that inputs two frames. The feature scales $L=4$. For the data augmentation, the input RS frames are first randomly cropped with a width of $256$ while keeping the height unchanged, and a random horizontal flip is performed on the cropped patch. The loss hyper-parameters are set to $\lambda_1=0.005$, $\lambda_2 = 0.05$. The model is trained for 150k iterations with a step learning rate adjustment strategy. When testing, no augmentation is applied to the input consecutive RS frames. The experiments are conducted on the PyTorch platform on a single NVIDIA V100 GPU. The initial learning rate is set to $4\times 10^{-4}$, and the ADAM optimizer~\cite{kingma2014adam} is employed to update the model parameters. 

\vspace{2mm}
\noindent\textbf{Evaluation metrics. }
Both PSNR and SSIM~\cite{zhou2004ssim} are employed to evaluate the correction accuracy quantitatively. Meanwhile, the learned perceptual metric LPIPS~\cite{zhang2018unreasonable} is also applied to measure the visual quality quantitatively. In addition, the corrected RS frames are also displayed for the qualitative comparison.

\begin{figure*}[!t]
    \centering
    \includegraphics[width=\linewidth]{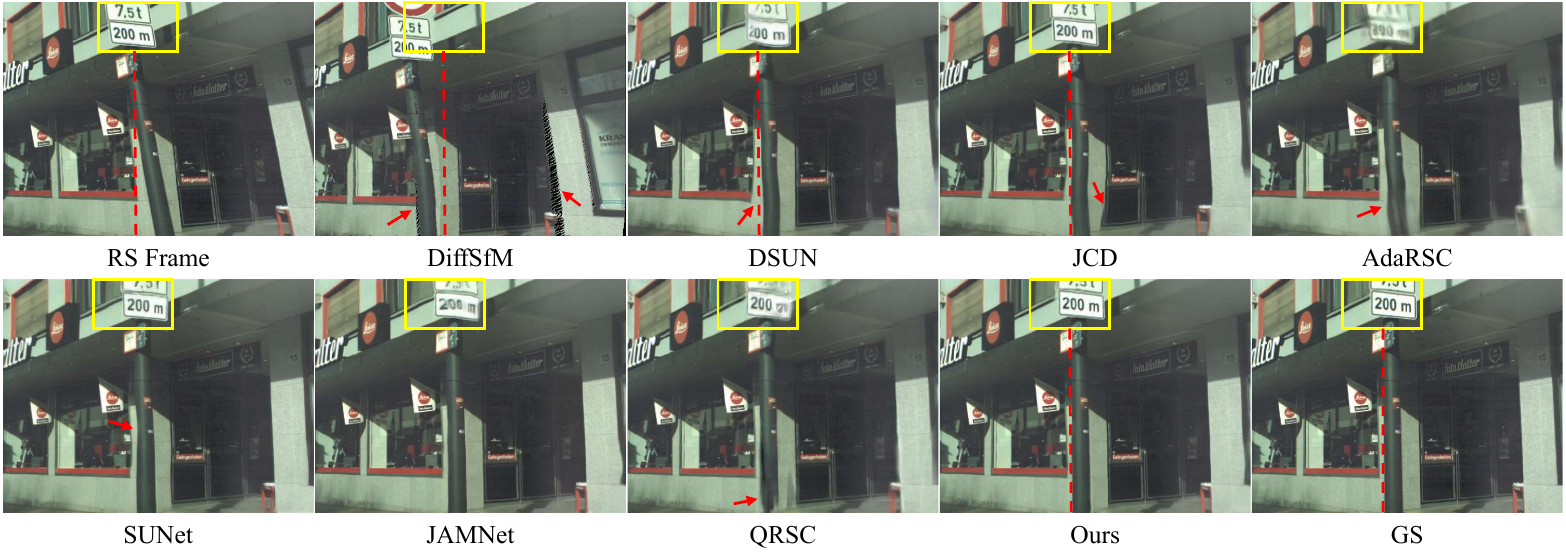}
    \caption{Qualitative results comparison against state-of-the-art methods on the synthetic Fastec-RS dataset~\cite{liu2020deep}. Our method removes the RS distortions well and preserves more details in the recovered GS image on such an occluded scene. }
    \label{fig:results_fastecrs}
\end{figure*}

\begin{figure*}[!t]
    \centering
    \includegraphics[width=\linewidth]{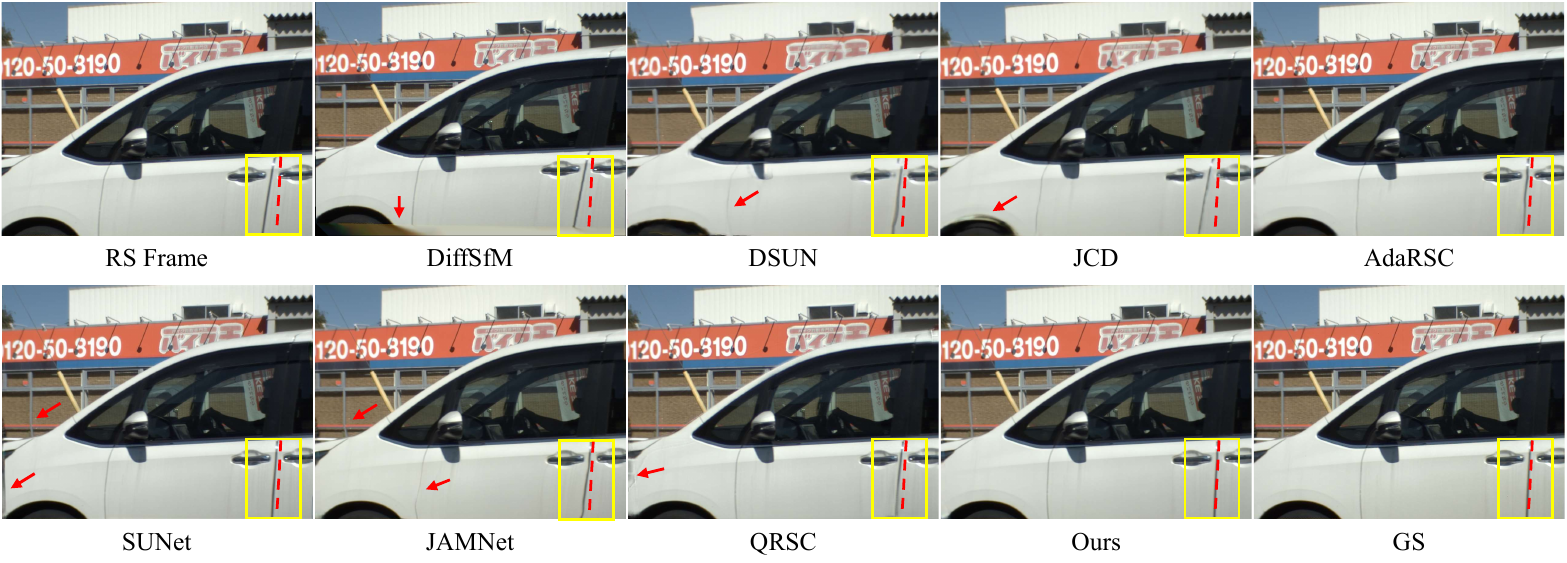}
    \caption{Qualitative results comparison against state-of-the-art methods on the real-world BS-RSC dataset~\cite{cao2022learning}. Our method is effective and robust in recovering the latent GS image accurately from the RS frames distorted by complex non-linear and large motions. }
    \label{fig:results_bsrsc}
\end{figure*}

\begin{table*}
    \centering
    \begin{subtable}[h]{0.5\linewidth}
    \centering
    \resizebox{\linewidth}{!}{
    \begin{tabular}{lccc}
    \toprule
    Model    &  PSNR & SSIM & \# Params. \\
    \midrule
    W/o motion modeling & 26.88 & 0.833 & 2.43 \\
    W/ undistortion flow $\textbf{U}_{r\rightarrow g}$ & 27.78 & 0.847 & 2.86 \\
    W/ distortion flow $\textbf{U}_{r\leftarrow g}$   & 28.39 & 0.864 & 2.79 \\
    \rowcolor{mygray} Full model & 28.88 & 0.870 & 2.87 \\
    \bottomrule
    \end{tabular}
    }
    \captionsetup{width=0.99\linewidth}
    \caption{Effectiveness of the distortion flow estimation. }
    \label{tab:ab_distortion_flow}
    \end{subtable}
    \hspace{3mm}
    \begin{subtable}[h]{0.45\linewidth}
    \centering
    \begin{tabular}{lccc}
    \toprule
    Model  &  PSNR & SSIM & \# Params. \\
    \midrule
    W/o Flow attention & 28.52 & 0.865 & 2.79 \\
    W/ 1 field  & 28.73 & 0.867 & 2.87 \\
    \rowcolor{mygray} W/ 4 fields &28.88 & 0.870 & 2.87 \\
    W/ 8 fields & 28.82 & 0.865 & 2.88 \\
    \bottomrule
    \end{tabular}
    \captionsetup{width=0.99\linewidth}
    \caption{Ablation study on the model design.}
    \label{tab:ab_model_design}
    \end{subtable}
    \vspace{-3mm}
    \caption{Ablation study of the motion modeling and the model design. The settings employed in our final model are highlighted.}
    \vspace{-5mm}
\end{table*}

\subsection{Comparison to the State-of-the-art}
We compare the proposed method to the state-of-the-art RSC methods quantitatively and qualitatively, including \textbf{1)} traditional method DiffSfM~\cite{zhuang2017rolling}, \textbf{2)} deep learning-based methods DSUN~\cite{liu2020deep}, SUNet~\cite{fan2021sunet}, VideoRS~\cite{naor2022combining}, JAMnet~\cite{fan2023joint} that take two consecutive frames as input, and \textbf{3)} deep learning-based methods JCD~\cite{zhong2021towards}, AdaRSC~\cite{cao2022learning}, QRSC~\cite{qu2023towards} that require inputting three or more frames. We also implemented two versions of the proposed method: 2-frame-based and 3-frame-based, to better demonstrate the effectiveness of the proposed method.

\vspace{1.5mm}
\noindent\textbf{Quantitative comparison. }
Table~\ref{tab:quantitative_results} presents the performance of different methods on the synthetic datasets Factec-RS and Carla-RS. We see that the proposed method achieved highly competitive performance that obtains higher PSNR, SSIM, and lower LPIPS than the state-of-the-art methods JAMNet~\cite{fan2023joint} and QRSC~\cite{qu2023towards}, thanks to the direct distortion flow estimation strategy and the model design. The quantitative results on the real-world BS-RSC dataset are shown in Tab.~\ref{tab:quantitative_results_real}, where the proposed method shows significant improvements against other methods. Specifically, our 3-frame-based model achieves 2.5dB PSNR improvement compared to the 3-frame-based QRSC (3F), and even surpasses the 5-frame-based QRSC (5F) with about 1dB PSNR. Note that BS-RSC contains both camera and object motions in the real world. 

These competitive results demonstrate the effectiveness of the proposed method in removing the RS effects under non-linear and large motions. Unlike previous methods that estimate the optical flows between RS frames and utilize linear~\cite{liu2020deep, cao2022learning, fan2023joint} or quadratic~\cite{qu2023towards} motion models to obtain the correction fields, our methods directly predict the distortion flow and can better model the non-linear motions to obtain better results. Meanwhile, as shown in the right part of Tab.~\ref{tab:quantitative_results_real}, our method also achieved superior results on the ACC dataset~\cite{qu2023towards}, which is derived from BS-RSC by excluding frames with constant motions. These quantitative results demonstrate that the proposed method is effective and robust in removing RS distortions under complex non-linear and large motions.

\vspace{1mm}
\noindent\textbf{Qualitative comparison. }
Figures~\ref{fig:results_fastecrs} and~\ref{fig:results_bsrsc} illustrate the qualitative results of different methods on the synthetic dataset Fastec-RS and the real-world dataset BS-RSC, respectively. As for the occluded scene shown in Fig.~\ref{fig:results_fastecrs}, we see that existing methods struggle to either remove the distortions or preserve details for high-quality GS restoration. In contrast, our method successfully recovers the corrected GS image while preserving more details (\eg, the number marked with the yellow box). Meanwhile, as for the scene containing both non-linear camera motion and object motion shown in Fig.~\ref{fig:results_bsrsc}, existing methods can hardly obtain the correct shape of the latent GS image (marked by the yellow box). These methods make it difficult to obtain an accurate correction field with a linear motion model, \eg, DSUN, AdaRSC, and JAMNet, even with a quadratic motion model, \ie, QRSC. Thanks to the proposed direct intermediate distortion flow estimation and the network design, our model performs better in removing the RS distortions caused by complex and large motions and recovering the desired GS image accurately. 

\vspace{1mm}
\noindent\textbf{Efficiency comparison. }
As shown in Tab.~\ref{tab:quantitative_results}, our method is also highly competitive in terms of efficiency. More specifically, our 2-frame-based model achieves a higher PSNR than the previous most efficient RSC model JAMNet, and has 40\% fewer parameters. Moreover, our method~(3-frame-based version) realizes a significant performance gain on all three datasets while only slightly slower than JAMNet. Compared to QRSC, the proposed method is more than 30 times faster with a much smaller number of model parameters. This is because QRSC requires computing the optical flows several times among input RS frames using the off-the-shelf flow models, while our method performs RSC in an end-to-end manner. 

\subsection{Ablation Studies}
We ablate the proposed method in terms of the distortion flow estimation and the network modules with our 2-frame-based model on the popular Fastec-RS dataset, and the ablation of our 3-frame-based model on the real-world BS-RSC dataset can be found in the supplementary materials.  

\vspace{1.5mm}
\noindent\textbf{Distortion flow estimation. }
To validate the effectiveness of the direct distortion flow estimation for the RSC task, we first remove the flow-attention module and multi-distortion flow decoding module, then \textbf{1)} remove the flow estimation and warping operation to obtain a vanilla encoder-decoder-like model (without motion modeling), \textbf{2)} replace the distortion flow estimation with direct undistortion flow $\mathbf{U}_{r \rightarrow g}$ estimation and apply differential forward warping~\cite{liu2020deep} module for warping. The results of the above model variants are shown in Tab.~\ref{tab:ab_distortion_flow}. We see that a vanilla encoder-decoder model achieves the lowest metrics, while the models with motion modeling~(with undistortion or distortion flow) significantly improve the performance. This verifies that inter-frame motion modeling is beneficial and necessary to achieve high-quality RSC results. Meanwhile, the model with distortion flow estimation obtains higher PSNR and SSIM than the undistortion flow-based model. We argue that the backward warping operation contributes to the performance improvement. In addition, when the distortion flow-based model with the delicate network design (\ie, the global correlation-based flow attention and multi-distortion flow decoding strategy), the performance has been further improved with a slight parameter number increase. 

\vspace{1.5mm}
\noindent\textbf{Flow attention and multi-flow decoder. }
As shown in Tab.~\ref{tab:ab_model_design}, when adding the flow attention module or the multi-distortion flow decoding, PSNR and SSIM metrics have been further improved by better modeling the large complex motions and occlusions. Meanwhile, as the flow group number increases, the model exhibits minor performance fluctuations. However, it still achieves improvement when compared to the single-field-based model.

\vspace{1.5mm}
\noindent\textbf{The Number of input RS frames. }
As shown in Tabs.~\ref{tab:quantitative_results} and~\ref{tab:quantitative_results_real}, the performance of the two-frame-based version model declines drastically on all datasets. With two frames, some contents in the latent GS image corresponding to the middle scanline of the second RS image still cannot be found in the first RS image. As a result, the missing regions should be generated, which is highly challenging. When with three frames, more complementary appearance information in the neighboring RS frames can be aggregated to obtain a higher-quality GS image.
\section{Limitation}
Although our method achieves highly competitive performance in recovering high-quality GS images and surpasses existing methods with a large margin, it still can not fully address the RS distortions encountered in real-world scenarios, constrained by the scale of existing datasets and the unknown camera parameters of the capture. In the following work, we want to integrate explicit camera exposure parameters into the model design for a more effective and generalized real-world RSC task.

\section{Conclusion}
This paper explores the intermediate distortion flow estimation for the high-quality performance on the RSC task. A novel framework, equipped with a global correlation-based flow attention module and a multi-distortion flow decoding strategy, is proposed to estimate the distortion flows from the latent GS image to the RS frames directly. Experimental results on both synthetic and real-world datasets demonstrate the effectiveness of the proposed method, and that it can remove the RS distortions under complex non-linear and large motions efficiently. We hope the proposed method can serve as a new paradigm to develop more effective and efficient methods for the RSC task.

\vspace{1.6mm}
\noindent\textbf{Acknowledgements}.
This research was supported in part by JSPS KAKENHI Grant Numbers 22H00529, 20H05951, JST-Mirai Program JPMJMI23G1. 

\newpage
{
    \small
    \bibliographystyle{ieeenat_fullname}
    \bibliography{main}
}


\end{document}